\documentclass[10pt,twocolumn,letterpaper]{article}

\usepackage{cvpr}
\usepackage{times}
\usepackage{epsfig}
\usepackage{graphicx}
\usepackage{amsmath}
\usepackage{amssymb}

\usepackage{subfig}
\usepackage{multirow}
\usepackage[bookmarks=true,bookmarksnumbered=true,
bookmarksopen=false,colorlinks=true]{hyperref} 

\newcommand{\T}{^\mathsf{T}}

\cvprfinalcopy 

\begin{document}

\title{GMM-Based Hidden Markov Random Field\\ for Color Image and 3D Volume Segmentation}

\author{Quan Wang\\
Signal Analysis and Machine Perception Laboratory\\
Electrical, Computer, and Systems Engineering\\
Rensselaer Polytechnic Institute\\
{\tt\small wangq10@rpi.edu}
}

\maketitle

\begin{abstract}
In this project\footnote{This work originally appears as the final project of 
Prof. \href{http://www.ecse.rpi.edu/~qji/}{Qiang Ji}'s course \emph{Introduction to Probabilistic Graphical Models} at RPI.}, we first study the Gaussian-based hidden Markov random field (HMRF) model and its expectation-maximization (EM) algorithm. Then we generalize it to Gaussian mixture model-based hidden Markov random field. 
The algorithm is implemented in {\tt MATLAB}. We also apply this algorithm to color image segmentation problems and 3D volume segmentation problems. 
\end{abstract}

\section{Introduction}
Markov random fields (MRFs) have been widely used for computer vision problems, such as
image segmentation \cite{zhanglei}, surface reconstruction \cite{surface} 
and depth inference \cite{depth}. Much of its success attributes to the efficient algorithms, 
such as Iterated Conditional Modes \cite{ICM}, and its consideration of both 
``data faithfulness'' and ``model smoothness'' \cite{AGSM}. 

The HMRF-EM framework was first proposed for segmentation of brain MR images
\cite{HMRF-EM}. 
For simplicity, we first assume that the image is 2D gray-level, 
and the intensity distribution of each region to be segmented follows a Gaussian distribution. 
Given an image $\textbf{Y}=(y_1,\dots,y_N)$ where $N$ is the number of pixels and 
each $y_i$ is the gray-level intensity of a pixel, 
we want to infer a configuration of labels $\textbf{X}=(x_1,\dots,x_N)$ where $x_i \in L$ and $L$
is the set of all possible labels. In a binary segmentation
problem, $L= \lbrace 0,1 \rbrace$. According to the MAP 
criterion, we seek the labeling $\textbf{X}^\star$ which satisfies:
\begin{equation}
\label{eq:MAP}
\textbf{X}^\star=\underset{\textbf{X}}{\operatorname{argmax}} \; \lbrace 
P(\textbf{Y}|\textbf{X},\Theta)P(\textbf{X}) \rbrace  .
\end{equation}
The prior probability $P(\textbf{X})$ is a Gibbs distribution, and the joint likelihood
probability is
\begin{eqnarray}
P(\textbf{Y}|\textbf{X},\Theta)&=&\prod\limits_i P(y_i|\textbf{X},\Theta) \nonumber \\ 
&=&\prod\limits_i P(y_i|x_i,\theta_{x_i}) ,
\end{eqnarray}
where $P(y_i|x_i,\theta_{x_i})$ is a Gaussian distribution with parameters
$\theta_{x_i}=(\mu_{x_i},\sigma_{x_i})$. 
In MRF problems, people usually learn the parameter set 
$\Theta=\lbrace \theta_l | l \in L \rbrace$ from the training data. 
For example, in image segmentation problems, prior knowledge of the intensity distributions of 
the foreground and the background might be consistent within a dataset, especially domain specific dataset. 
Thus, we can learn the 
parameters from some images that are manually labeled, and use these parameters to run the MRF 
to segment the other images. 

The major difference between MRF and HMRF is that, in HMRF, the parameter set $\Theta$ is learned 
in an unsupervised manner. In a HMRF image segmentation problem, there is no training stage, and we 
assume no prior knowledge is known about the foreground/background intensity distribution. 
Thus, a natural proposal for solving a HMRF problem is to use the EM algorithm, 
where parameter set $\Theta$ and label configuration $\textbf{X}$ are learned alternatively. 

\section{EM Algorithm for Parameters}
\label{sec:EM}
We still use the 2D gray-level and Gaussian distribution assumption. 
We use the EM algorithm to estimate the parameter set 
$\Theta=\lbrace \theta_l | l \in L \rbrace$. 
We describe the EM algorithm by the following \cite{hmrf-em-image}:
\begin{enumerate}
\item
\textit{Start:} Assume we have an initial parameter set $\Theta^{(0)}$.

\item
\textit{E-step:} At the $t$th iteration, we have $\Theta^{(t)}$, and 
we calculate the conditional expectation:
\begin{eqnarray}
\hspace{-5mm}
Q(\Theta | \Theta^{(t)}) &=& E\left[ 
\ln P(\textbf{X},\textbf{Y}|\Theta) | \textbf{Y},\Theta^{(t)}
\right]
\nonumber \\
&=& \sum\limits_{\textbf{X}\in \chi}
P(\textbf{X}|\textbf{Y},\Theta^{(t)}) \ln P(\textbf{X},\textbf{Y}|\Theta) ,
\end{eqnarray}
where $\chi$ is the set of all possible configurations of labels. 

\item
\textit{M-step:} Now maximize $Q(\Theta | \Theta^{(t)})$
to obtain the next estimate:
\begin{equation}
\Theta^{(t+1)}=\underset{\Theta}{\operatorname{argmax}} \;  Q(\Theta | \Theta^{(t)}) .
\end{equation}
Then let $\Theta^{(t+1)} \rightarrow \Theta^{(t)}$ and repeat from the E-step. 
\end{enumerate}

Let $G(z;\theta_l)$ denote a Gaussian distribution function with parameters 
$\theta_l=(\mu_l,\sigma_l)$:
\begin{equation}
\label{eq:gaussian}
G(z;\theta_l)=\dfrac{1}{\sqrt{2\pi\sigma_l^2}} \exp\left(
-\dfrac{(z-\mu_l)^2}{2\sigma_l^2}
\right) .
\end{equation}
We assume that the prior probability can be written as 
\begin{eqnarray}
P(\textbf{X})=\dfrac{1}{Z}\exp\left( -U(\textbf{X}) \right) ,
\end{eqnarray}
where $U(\textbf{x})$ is the prior energy function. 
We also assume that 
\begin{eqnarray}
P(\textbf{Y}|\textbf{X},\Theta)&=&
\prod\limits_i P(y_i|x_i,\theta_{x_i}) \nonumber \\
&=& \prod\limits_i G(y_i;\theta_{x_i}) \nonumber \\
&=& \dfrac{1}{Z'}\exp\left( -U(\textbf{Y}|\textbf{X}) \right) .
\end{eqnarray}
With these assumptions, the HMRF-EM algorithm is given below:
\begin{enumerate}
\item 
Start with initial parameter set $\Theta^{(0)}$.

\item
Calculate the likelihood distribution $P^{(t)}(y_i|x_i,\theta_{x_i})$.

\item
Using current parameter set $\Theta^{(t)}$ to estimate the labels by
MAP estimation:
\begin{eqnarray}
\textbf{X}^{(t)}&=&\underset{\textbf{X}\in \chi}{\operatorname{argmax}} \; \lbrace 
P(\textbf{Y}|\textbf{X},\Theta^{(t)})P(\textbf{X}) \rbrace \nonumber \\
&=&\underset{\textbf{X}\in \chi}{\operatorname{argmin}} \; \lbrace 
U(\textbf{Y}|\textbf{X},\Theta^{(t)})+U(\textbf{X}) \rbrace .
\end{eqnarray}
The algorithm for the MAP estimation is discussed in Section \ref{section:MAP}. 

\item
Calculate the posterior distribution for all $l\in L$ and all pixels $y_i$ using the Bayesian rule:
\begin{eqnarray}
P^{(t)}(l | y_i)=\dfrac{G(y_i;\theta_l)P(l | x_{N_i}^{(t)})}{P^{(t)}(y_i)} ,
\end{eqnarray}
where $x_{N_i}^{(t)}$ is the neighborhood configuration of $x_i^{(t)}$, and
\begin{equation}
P^{(t)}(y_i)=\sum\limits_{l\in L}G(y_i;\theta_l)P(l | x_{N_i}^{(t)}) .
\end{equation}
Note here we have
\begin{eqnarray}
P(l | x_{N_i}^{(t)}) &=& 
\dfrac{1}{Z}\exp \left( -\sum\limits_{j\in N_i} V_c(l,x_j^{(t)})\right) .
\end{eqnarray}

\item
Use $P^{(t)}(l | y_i)$ to update the parameters:
\begin{eqnarray}
\mu_l^{(t+1)}&=&\dfrac{\sum\limits_{i}P^{(t)}(l|y_i)y_i}{\sum\limits_{i}P^{(t)}(l|y_i)} \\
(\sigma_l^{(t+1)})^2&=&
\dfrac{\sum\limits_{i}P^{(t)}(l|y_i)(y_i-\mu_l^{(t+1)})^2}{\sum\limits_{i}P^{(t)}(l|y_i)}  .
\end{eqnarray}
\end{enumerate}

\section{MAP Estimation for Labels}
\label{section:MAP}
In the EM algorithm, we need to solve for $\textbf{X}^\star$ that minimizes the total 
posterior energy
\begin{equation}
\label{eq:MAP2}
\textbf{X}^{\star}
=\underset{\textbf{X}\in \chi}{\operatorname{argmin}} \; \lbrace 
U(\textbf{Y}|\textbf{X},\Theta)+U(\textbf{X}) \rbrace 
\end{equation}
with given $\textbf{Y}$ and $\Theta$, where the likelihood energy (also called unitary potential) is
\begin{eqnarray}
\label{eq:like_energy}
U(\textbf{Y}|\textbf{X},\Theta)&=&
\sum\limits_{i} U(y_i|x_i,\Theta) \nonumber \\
&=& \sum\limits_{i} \left[
\dfrac{(y_i-\mu_{x_i})^2}{2\sigma_{x_i}^2}+\ln \sigma_{x_i} \right] .
\end{eqnarray}
The prior energy function (also called pairwise potential) $U(\textbf{X})$ has the form
\begin{equation}
U(\textbf{X})=\sum\limits_{c\in C} V_c(\textbf{X}) ,
\end{equation}
where $V_c(\textbf{X})$ is the clique potential and $C$ is the set of all possible cliques. 

In the image domain, we assume that one pixel has at most 4 neighbors: the pixels in 
its 4-neighborhood. Then the clique potential is defined on pairs of neighboring pixels:
\begin{eqnarray}
\label{eq:Vc}
V_c(x_i,x_j)=\dfrac{1}{2}(1-I_{x_i,x_j}) ,
\end{eqnarray}
where
\begin{equation}
I_{x_i,x_j}=
\left\{\begin{array}{c}
0 \qquad \textrm{if $x_i \neq x_j$}\\ 
1 \qquad \textrm{if $x_i = x_j$}
\end{array}\right. .
\end{equation}
Note that in Eq. (\ref{eq:Vc}), the constant coefficient $1/2$ can be replaced by a variable 
coefficient $\beta$. We just follow \cite{HMRF-EM} to use the $1/2$ constant, which proves effective in many of 
our experiment results. 

We developed an iterative algorithm to solve (\ref{eq:MAP2}):
\begin{enumerate}
\item
To start with, we have an initial estimate $\textbf{X}^{(0)}$, which can be from the
previous loop of the EM algorithm. 

\item 
\label{item:MAP_iter}
Provided $\textbf{X}^{(k)}$, for all $1\leq i\leq N$, we find
\begin{equation}
\label{eq:MAP_iter}
x_i^{(k+1)}=
\underset{l \in L}{\operatorname{argmin}} \; \lbrace
U(y_i | l)+\sum\limits_{j\in N_i} V_c(l,x_j^{(k)})
\rbrace .
\end{equation}

\item
Repeat step \ref{item:MAP_iter} until $U(\textbf{Y}|\textbf{X},\Theta)+U(\textbf{X})$ stops
changing significantly or a maximum $k$ is achieved. 
\end{enumerate}

\section{GMM-Based HMRF}
In previous sections, we have been assuming that the intensity distribution of each region to be segmented 
follows a Gaussian distribution with parameters $\theta_{x_i}=(\mu_{x_i},\sigma_{x_i})$. 
However, this is a very strong hypothesis which is insufficient to model the complexity of the intensity distribution of real-life objects, especially for objects with multimodal distributions. 

Gaussian mixture model (GMM), in contrast, is much more powerful for modeling the complex distributions than one single Gaussian distribution. A Gaussian mixture model with $g$ components can be represented by parameters:
 \begin{eqnarray}
 \label{eq:gmm_para}
 \theta_l=\lbrace (\mu_{l,1},\sigma_{l,1},w_{l,1}),\dots,(\mu_{l,g},\sigma_{l,g},w_{l,g})\rbrace  .
 \end{eqnarray}
Compared with Eq. (\ref{eq:gaussian}), the GMM now has a weighted probability
\begin{eqnarray}
G_{\textrm{mix}}(z;\theta_l)=\sum\limits_{c=1}^g w_{l,c}G(z;\mu_{l,c},\sigma_{l,c})  .
\end{eqnarray}

Now, the M-step of the EM-algorithm described in Section \ref{sec:EM} changes to 
a Gaussian mixture model fitting problem. The GMM fitting problem itself can be also solved 
using an EM-algorithm. In the E-step, we determine which data should belong to which Gaussian component; 
in the M-step, we recompute the GMM parameters.

\section{Experiment Results}
We use the above mentioned GMM-based HMRF for two applications: color image segmentation and 
3D volume segmentation. For each application, minor modifications need to be made. 

\subsection{Color Image Segmentation}

\begin{figure*}
  \centering
  \subfloat[]
  {\includegraphics[width=0.45\textwidth]{385028} }  
  \;    
  \subfloat[]
  {\includegraphics[width=0.45\textwidth]{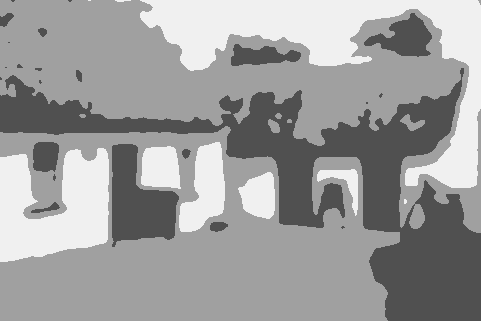} }
  \;      \\
  \subfloat[]
  {\includegraphics[width=0.45\textwidth]{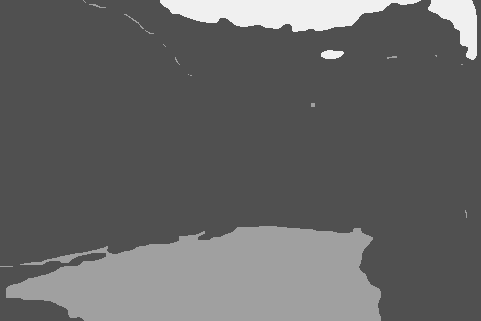} }
   \;      
  \subfloat[]
  {\includegraphics[width=0.46\textwidth]{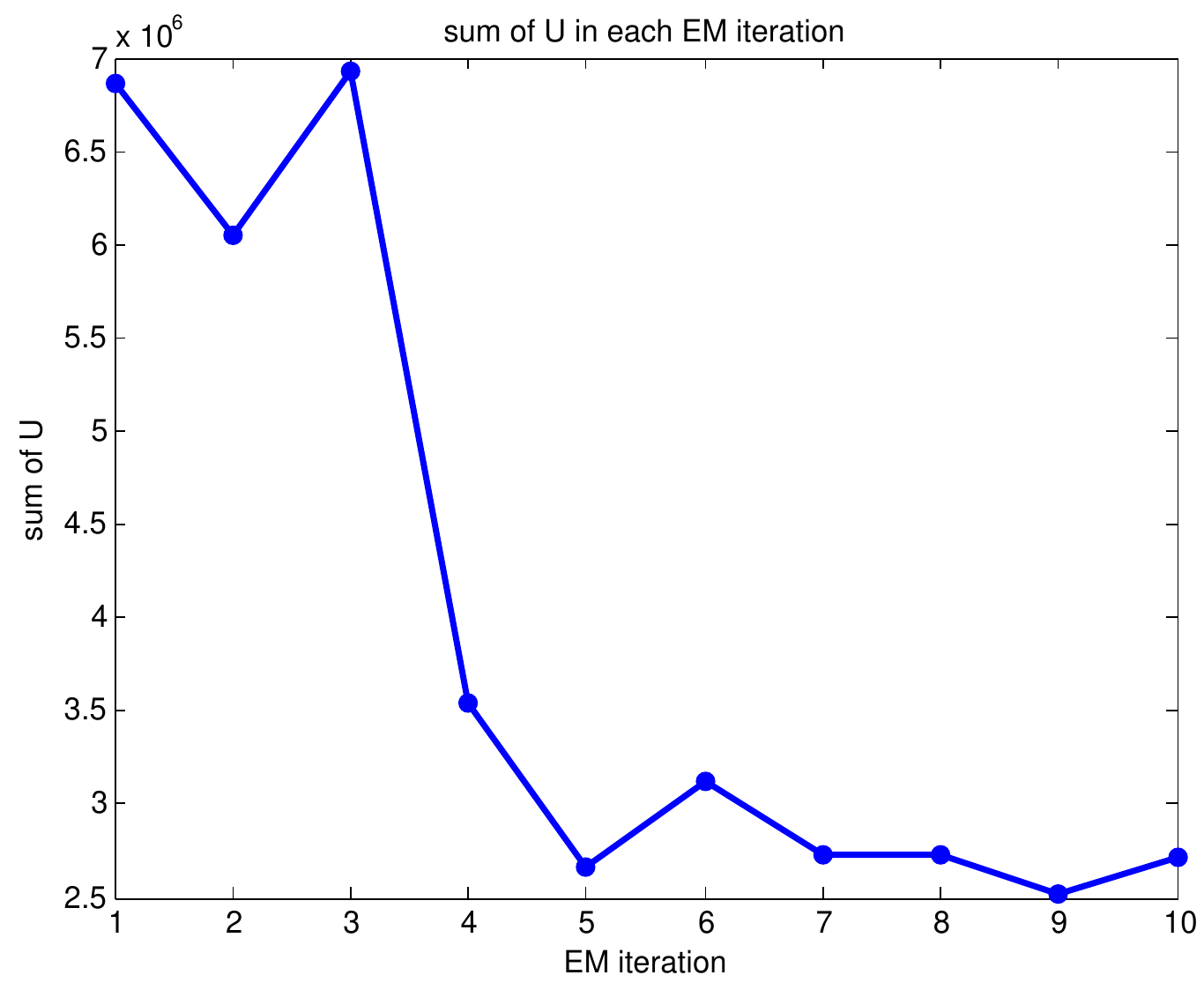} }
  \caption{Example color image segmentation results. 
  (a) Original color image. 
  (b) Initial segmentation by k-means.
  (c) Final segmentation by HMRF. 
  (d) Sum of energy in each iteration.
  }
  \label{fig:color_result_1}
\end{figure*}

\begin{figure*}
  \centering
  \subfloat[]
  {\includegraphics[width=0.25\textwidth]{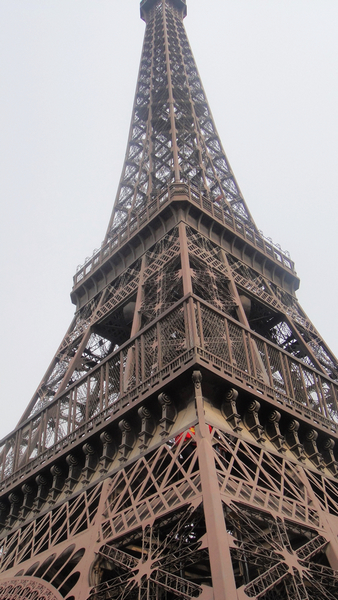} }  
  \;    
  \subfloat[]
  {\includegraphics[width=0.25\textwidth]{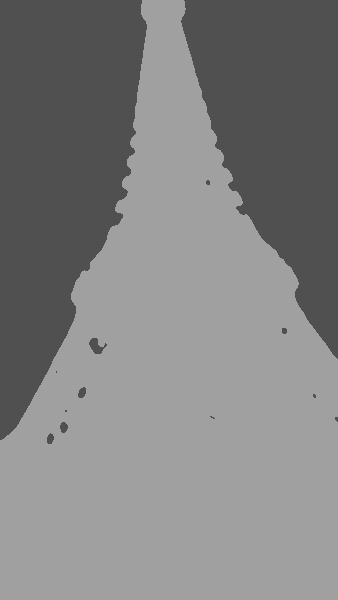} }
  \;     
  \subfloat[]
  {\includegraphics[width=0.25\textwidth]{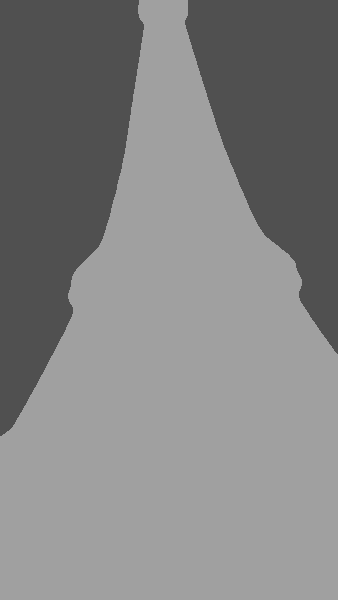} }
  \\
  \subfloat[]
  {\includegraphics[width=0.25\textwidth]{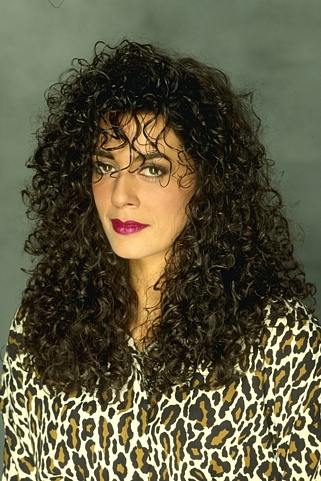} }  
  \;    
  \subfloat[]
  {\includegraphics[width=0.25\textwidth]{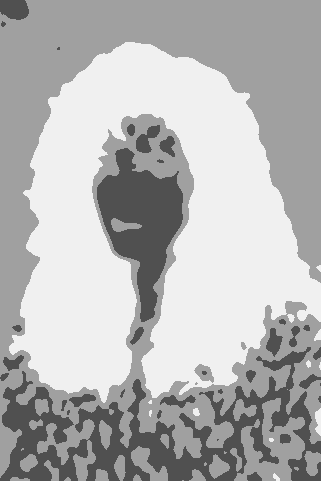} }
  \;     
  \subfloat[]
  {\includegraphics[width=0.25\textwidth]{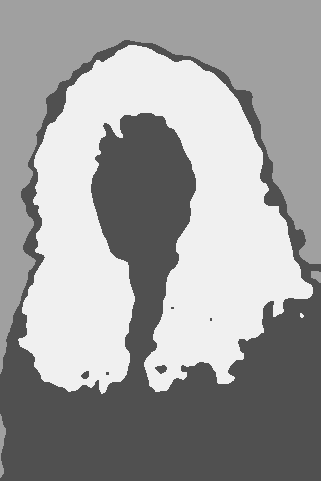} }
  \caption{More color image segmentation results. 
  First column: original color image;
  second column: initial segmentation by k-means;
  third column: final segmentation by HMRF. 
  }
  \label{fig:color_result_2}
\end{figure*}

\begin{figure*}
  \centering
  \subfloat[]
  {\includegraphics[width=0.25\textwidth]{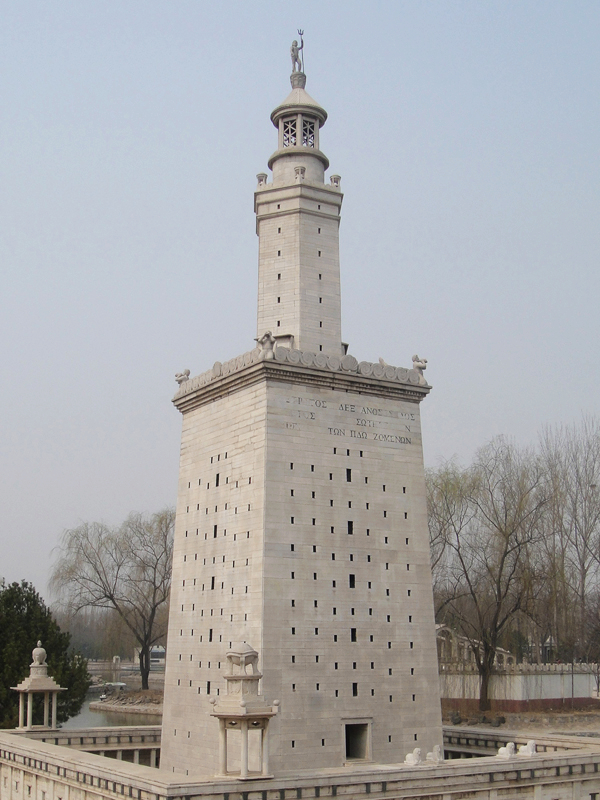} }  
  \;    
  \subfloat[]
  {\includegraphics[width=0.25\textwidth]{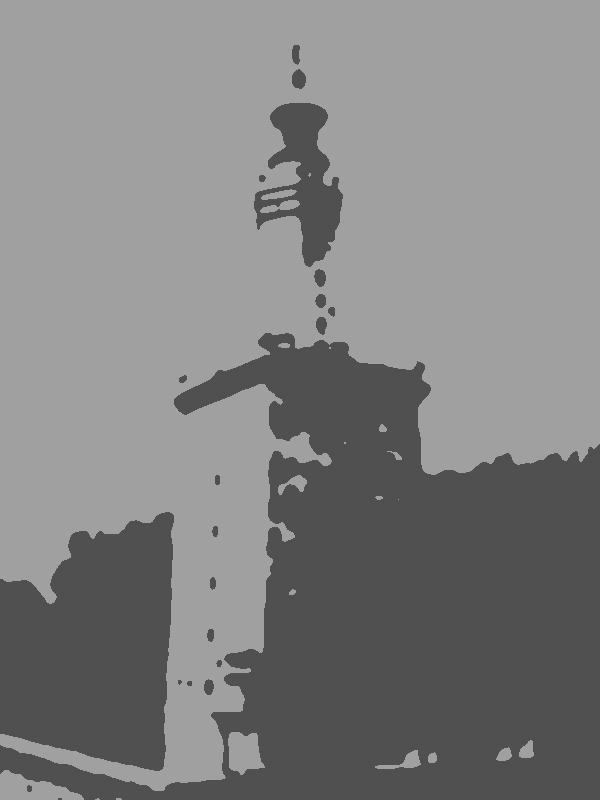} }
  \;     
  \subfloat[]
  {\includegraphics[width=0.25\textwidth]{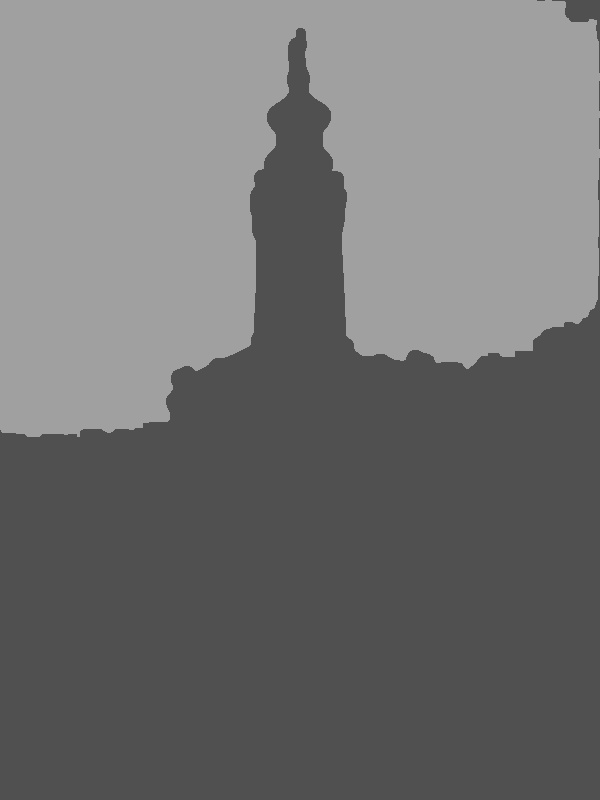} }
  \\
  \subfloat[]
  {\includegraphics[width=0.25\textwidth]{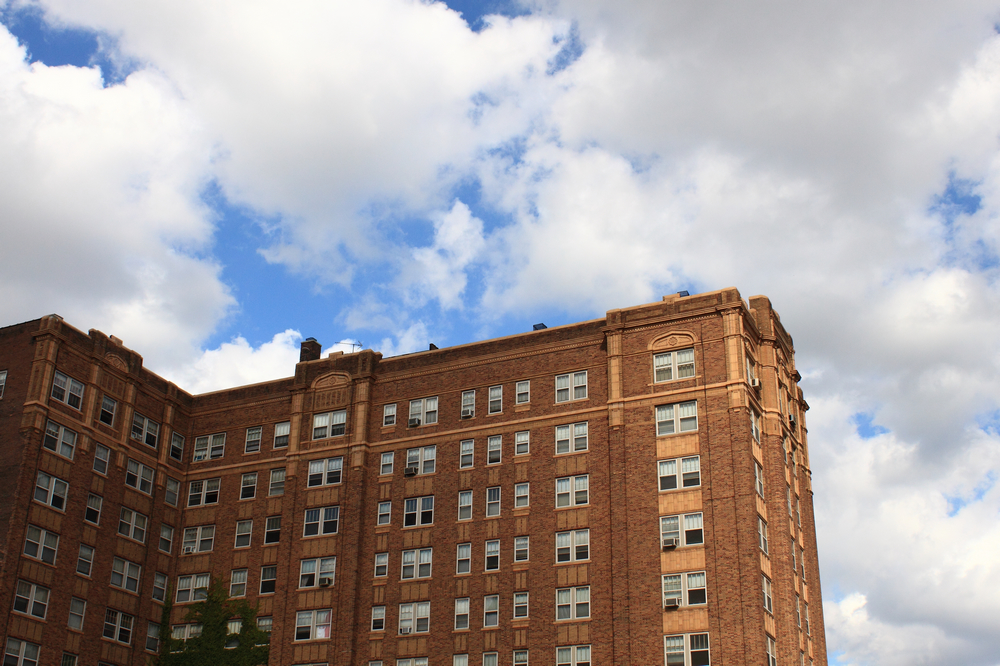} }  
  \;    
  \subfloat[]
  {\includegraphics[width=0.25\textwidth]{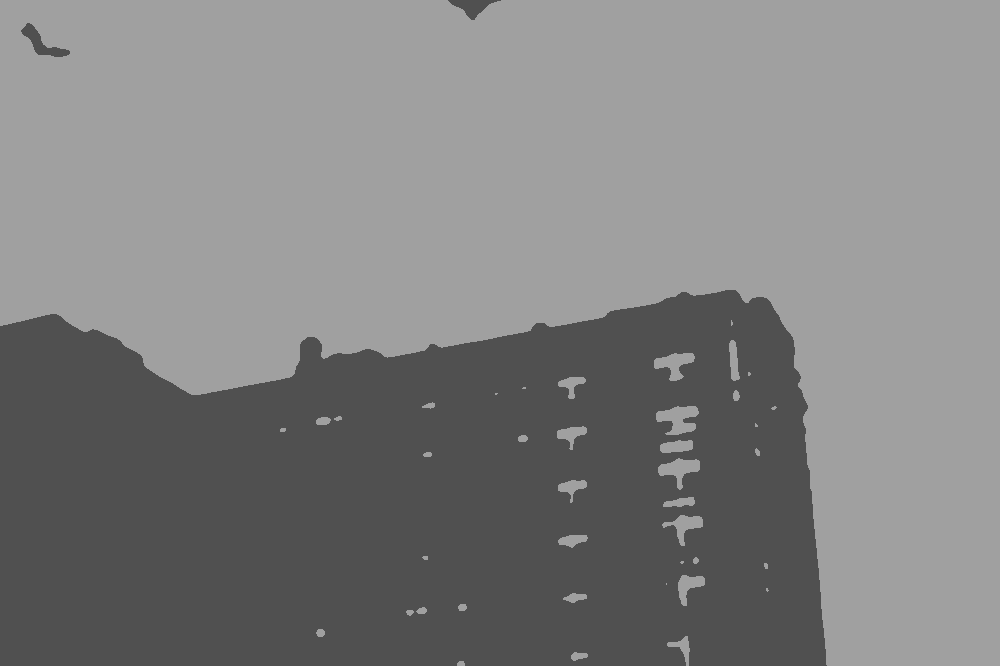} }
  \;     
  \subfloat[]
  {\includegraphics[width=0.25\textwidth]{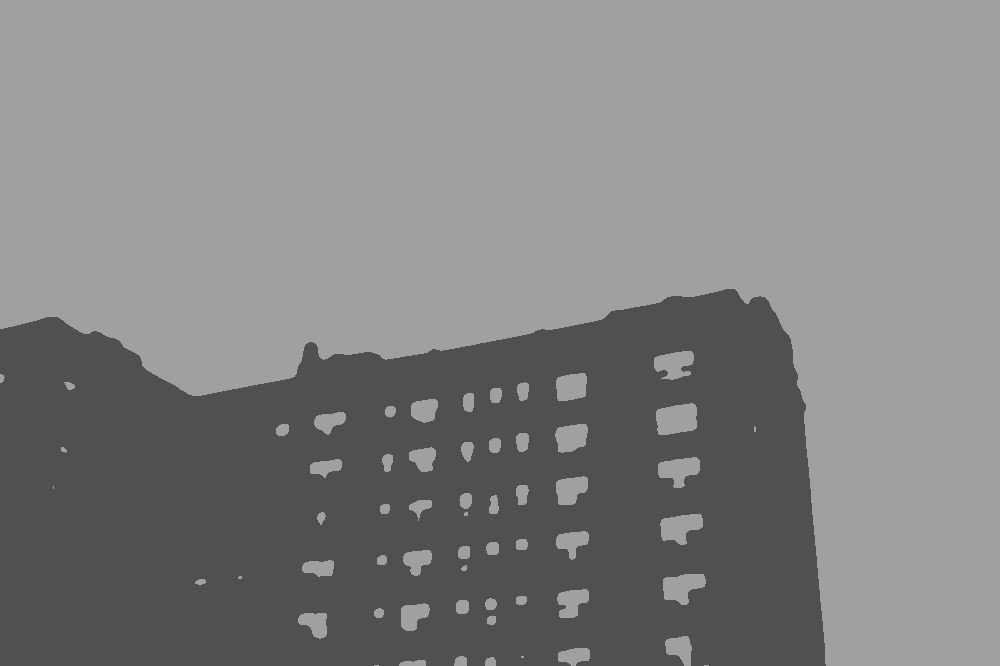} }
  \\
  \subfloat[]
  {\includegraphics[width=0.25\textwidth]{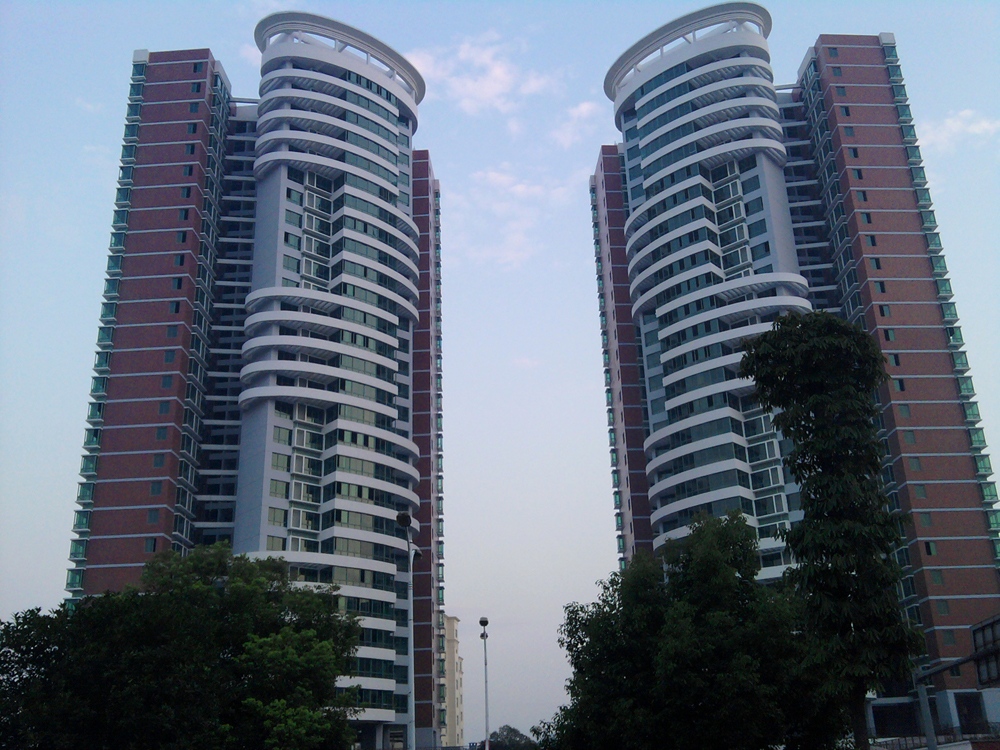} }  
  \;    
  \subfloat[]
  {\includegraphics[width=0.25\textwidth]{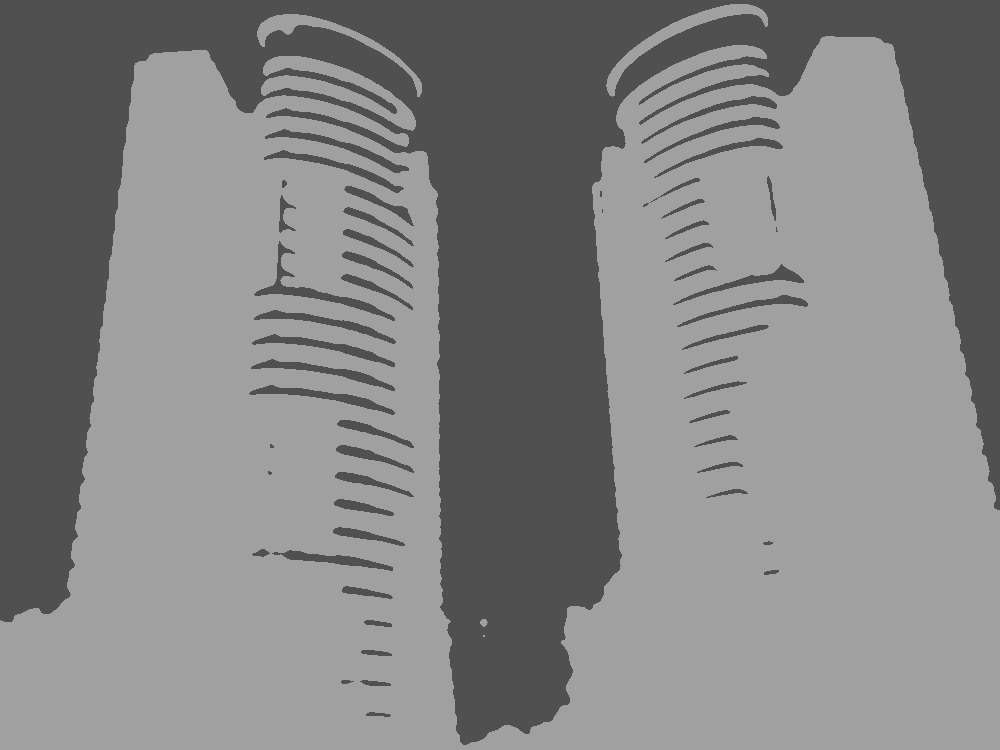} }
  \;      
  \subfloat[]
  {\includegraphics[width=0.25\textwidth]{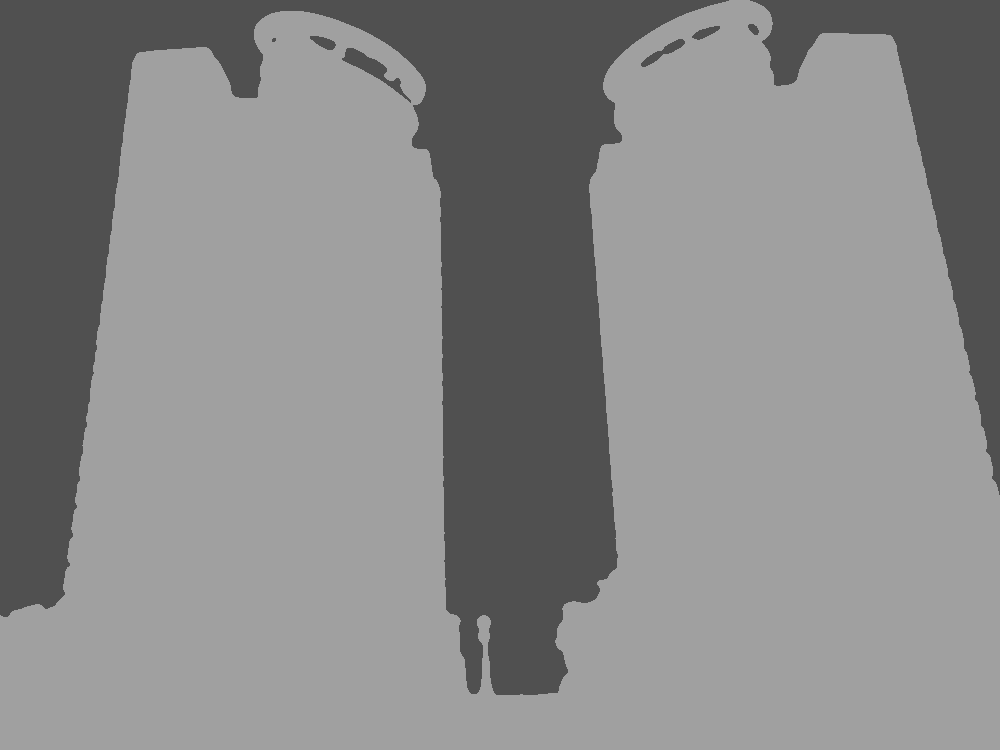} }
  \caption{More color image segmentation results (continued). 
  }
  \label{fig:color_result_3}
\end{figure*}

The difference between color image segmentation and gray-level image segmentation is that, for a color image, the pixel intensity is no longer a number, but a 3-dimensional vector of RGB values: $\textbf{Y}=(\textbf{y}_1,\dots,\textbf{y}_N)$, and $\textbf{y}_i=(y_{iR},y_{iG},y_{iB})\T$. The parameters of a Gaussian mixture model now becomes
 \begin{eqnarray}
 \label{eq:gmm_para_rgb}
 \theta_l=\lbrace (\boldsymbol\mu_{l,1},\boldsymbol\Sigma_{l,1},w_{l,1}),\dots,(\boldsymbol\mu_{l,g},\boldsymbol\Sigma_{l,g},w_{l,g})\rbrace ,
 \end{eqnarray}
which can be compared with Eq. (\ref{eq:gmm_para}). 
Also, the likelihood energy Eq. (\ref{eq:like_energy}) becomes
 \begin{eqnarray}
\label{eq:like_energy_rgb}
U(\textbf{Y}|\textbf{X},\Theta)&=&
\sum\limits_{i} U(\textbf{y}_i|x_i,\Theta) \nonumber \\
&=& \sum\limits_{i} \left[
\dfrac{1}{2}(\textbf{y}_i-\boldsymbol\mu_{x_i})\T \boldsymbol\Sigma_{x_i}^{-1} (\textbf{y}_i-\boldsymbol\mu_{x_i})
\right. \nonumber \\ 
&&+ \left. \ln |\boldsymbol\Sigma_{x_i}|^{\frac{1}{2}} \right] .
\end{eqnarray} 
Example color image segmentation results are shown in Figure \ref{fig:color_result_1}, \ref{fig:color_result_2}, and \ref{fig:color_result_3}.
 
\subsection{3D Volume Segmentation}

\begin{figure}
  \centering
\includegraphics[width=0.48\textwidth]{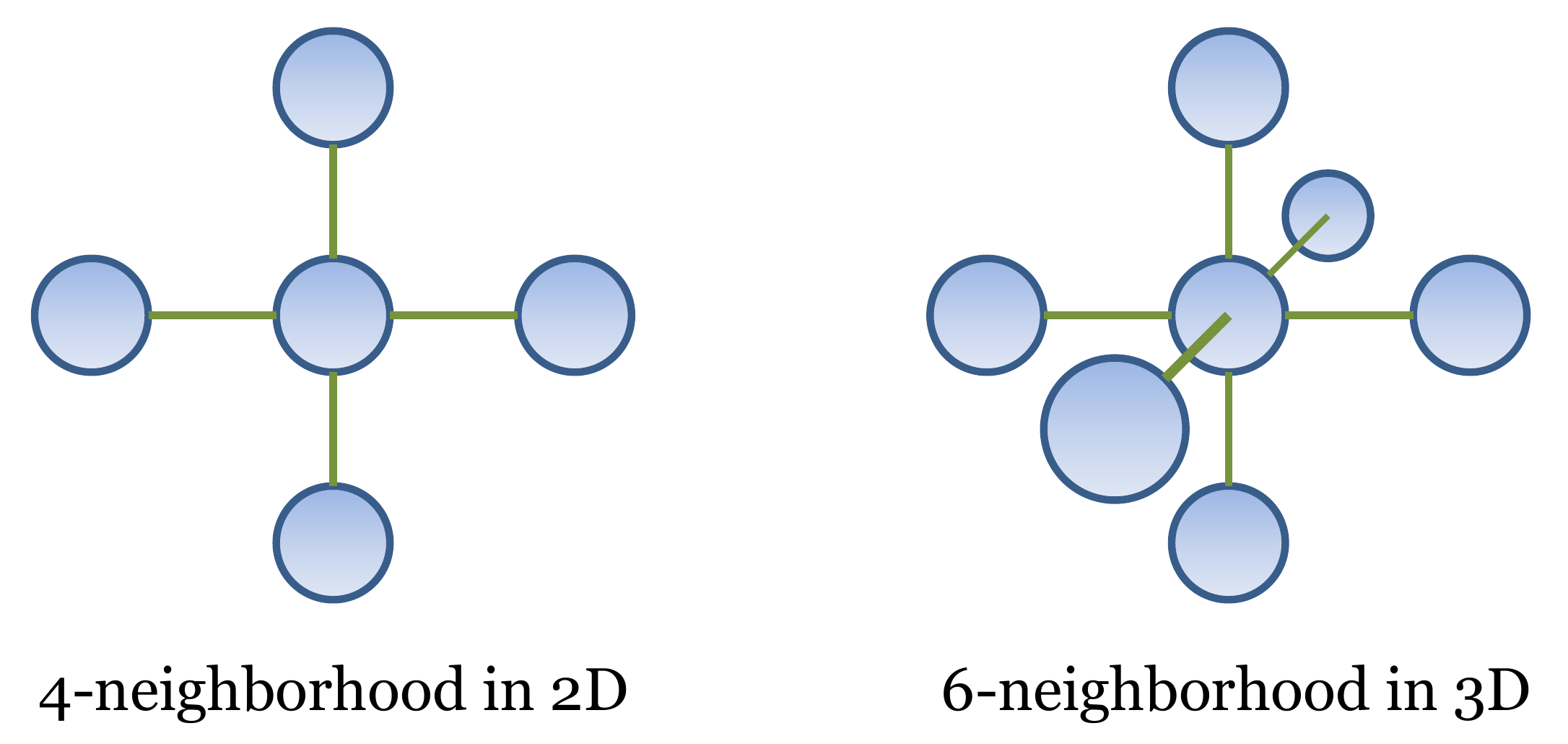} 
  \caption{Neighborhood system in 2D and 3D images.  
  }
  \label{fig:neighborhood}
\end{figure}

The only difference between 2D image segmentation and 3D image segmentation is the neighborhood system. 
In 2D images, we usually use the 4-neighborhood system or the 8-neighborhood system; in 3D images, we 
usually use the 6-neighborhood system or the 26-neighborhood system. The difference is shown in Figure 
\ref{fig:neighborhood}. 

\begin{figure*}
  \centering
  \subfloat[]
  {\includegraphics[height=0.4\textwidth]{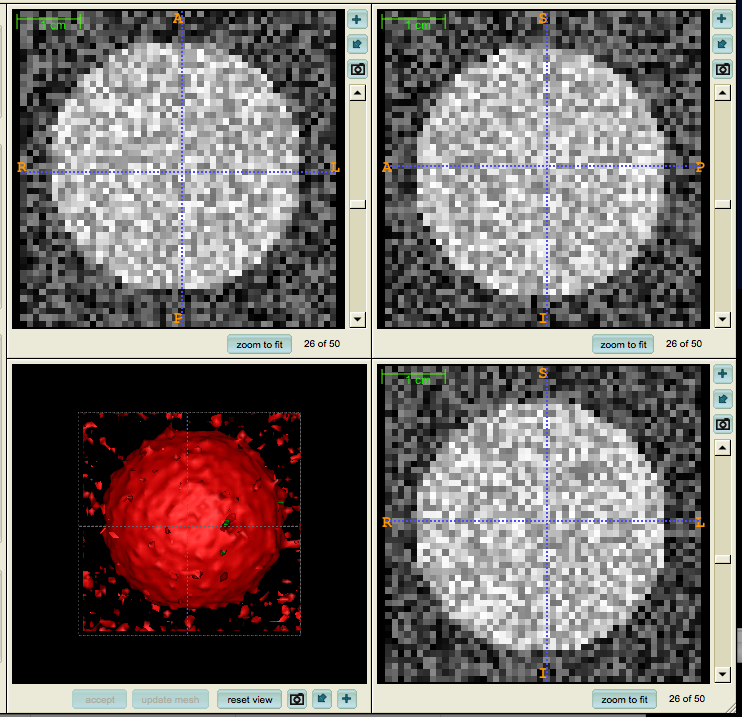} }  
  \quad
  \subfloat[]
  {\includegraphics[height=0.4\textwidth]{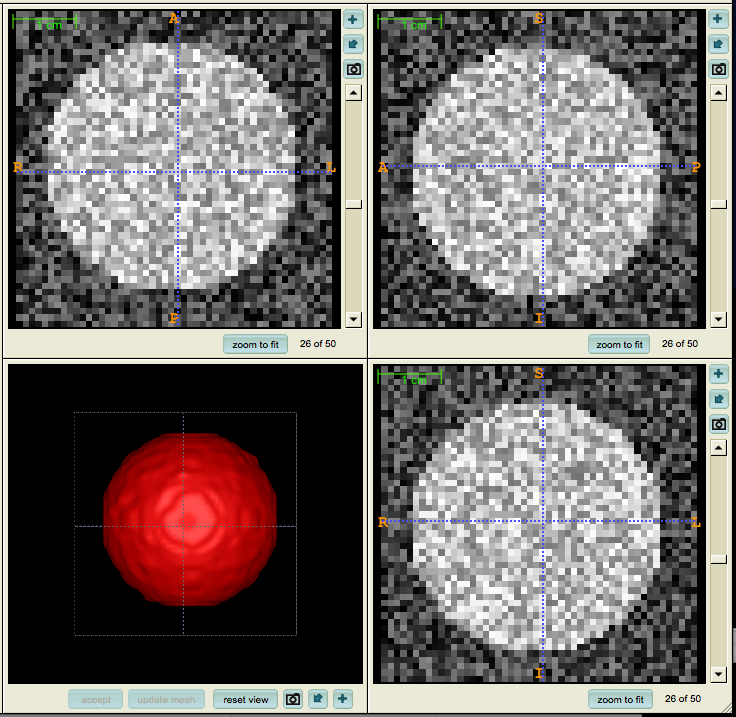} }      
  \caption{3D volume segmentation results: (a) k-means; (b) HMRF.  
  }
  \label{fig:3d_result}
\end{figure*}

To validate our algorithm for 3D volume segmentation, we generate a synthetic 3D image of size 
$50\times50\times50$ and with a foreground sphere of radius $20$ at the center. The intensity of background is $0$, 
and the foreground is $100$. Random noise uniformly distributed within $[0,120]$ is added to the entire image, at all positions. Thus 
clustering methods such as k-means will not guarantee spatial continuousness of the segmentation results. A comparison of k-means segmentation and HMRF segementation is shown in Figure \ref{fig:3d_result}. 
With 10 EM iterations and 10 MAP iterations, while setting $g=1$ for GMM, the 3D segmentation takes about 14 seconds 
on a 2.53 GHz Intel(R) Core(TM) i5 CPU.

\section{Code Documentation}
We provide the name and usage of each file of our  {\tt MATLAB} implementation in Tabel \ref{table:code1} and \ref{table:code2}, including both color image segmentation and 3D volume segmentation. 

\begin{table*}
\begin{center}
\begin{tabular}{|c|c|c|}
\hline
File & Type &
Usage
\\ \hline
\multirow{2}{*}{{\tt demo.m}} & \multirow{2}{*}{Runnable script} &
A color image segmentation example. \\ & & Users can run this file directly. \\ \hline
\multirow{2}{*}{{\tt image\_kmeans.m}} & \multirow{2}{*}{Function} &
The k-means algorithm for 2D color images. \\ & &
This will generate an initial segmentation. \\ \hline
{\tt HMRF\_EM.m} & Function & The HMRF-EM algorithm. \\ \hline
{\tt MRF\_MAP.m} & Function & The MAP algorithm. \\ \hline
{\tt gaussianBlur.m} & Function & Blurring an image using Gaussian kernel. \\ \hline
{\tt gaussianMask.m} & Function & Obtaining the mask of Gaussian kernel. \\ \hline
{\tt ind2ij.m} & Function & Index to 2D image coordinates conversion. \\ \hline
{\tt get\_GMM.m} & Function & Fitting Gaussian mixture model to data. \\ \hline
{\tt BoundMirrorExpand.m} & Function & Expanding an image. \\ \hline
{\tt BoundMirrorShrink.m} & Function & Shrinking an image. \\ \hline
{\tt 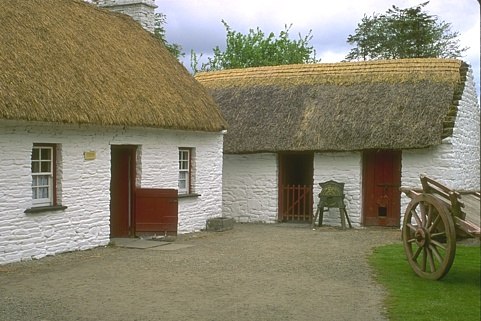} & Image & An example input color image.
\\ \hline
\end{tabular}
\end{center}
\caption{Name and usage of each file in color image segmentation {\tt MATLAB} code.}
\label{table:code1}
\end{table*}

\begin{table*}
\begin{center}
\begin{tabular}{|c|c|c|}
\hline
File & Type &
Usage
\\ \hline
\multirow{2}{*}{{\tt demo.m}} & \multirow{2}{*}{Runnable script} &
A 3D volume segmentation example. \\ & & Users can run this file directly. \\ \hline
{\tt generate\_3D\_image.m} & Runnable script & Generating the synthetic input 3D image. \\ \hline
\multirow{2}{*}{{\tt image\_kmeans.m}} & \multirow{2}{*}{Function} &
The k-means algorithm for 3D volumes. \\ & &
This will generate an initial segmentation. \\ \hline
{\tt HMRF\_EM.m} & Function & The HMRF-EM algorithm. \\ \hline
{\tt MRF\_MAP.m} & Function & The MAP algorithm. \\ \hline
{\tt ind2ijq.m} & Function & Index to 3D image coordinates conversion. \\ \hline
{\tt get\_GMM.m} & Function & Fitting Gaussian mixture model to data. \\ \hline
{\tt Image.raw} & Raw 3D image & An example input raw 3D image.
\\ \hline
\end{tabular}
\end{center}
\caption{Name and usage of each file in 3D volume segmentation {\tt MATLAB} code. }
\label{table:code2}
\end{table*}

\section{Discussion}
In this project, we have studied the hidden Markov random field, and its expectation-maximization algorithm. 
The basic idea of HMRF is combining ``data faithfulness'' and ``model smoothness'', which is very similar to 
active contours \cite{snakes}, gradient vector flow (GVF) \cite{gvf}, graph cuts \cite{gc}, and random walks \cite{rw}. 
We also combined the HMRF-EM framework with Gaussian mixture models, and applied it to 
color image segmentation and 
3D volume segmentation problems. 
The algorithms are implemented in {\tt MATLAB}.
In color image segmentation experiments, we can see the HMRF segmentation results are much more smooth than 
the results of direct k-means clustering. In 3D volume segmentation results, the segmented object is much closer to 
the original shape than clustering. This is because Markov random field imposes strong spatial constraints on 
the segmented regions, while clustering-based segmentation only considers pixel/voxel intensities.

{\small
\bibliographystyle{ieee}
\bibliography{egbib}
}

\end{document}